\documentclass[sigconf, nonacm]{acmart}
\newcommand\vldbdoi{10.14778/3617838.3617842}
\newcommand\vldbpages{41 - 50}
\newcommand\vldbvolume{17}
\newcommand\vldbissue{1}
\newcommand\vldbyear{2023}
\newcommand\vldbauthors{\authors}
\newcommand\vldbtitle{\shorttitle} 
\newcommand\vldbavailabilityurl{https://github.com/xkLi-Allen/FedGTA}
\newcommand\vldbpagestyle{empty}

\usepackage{graphicx}
\usepackage{textcomp}
\usepackage{xcolor}
\def\BibTeX{{\rm B\kern-.05em{\sc i\kern-.025em b}\kern-.08em
    T\kern-.1667em\lower.7ex\hbox{E}\kern-.125emX}}
\usepackage{balance} 
\usepackage{bbm}
\usepackage{amsfonts}
\usepackage{epstopdf}
\usepackage{array}
\usepackage{booktabs}
\usepackage{color}
\usepackage{xcolor}
\usepackage{colortbl}
\usepackage{xspace}
\usepackage{multirow}
\usepackage{pifont}
\usepackage{enumitem}
\usepackage{bbding}
\usepackage{fontawesome}
\usepackage{hyperref}
\usepackage{makecell}
\usepackage{microtype}
\usepackage{caption,setspace}
\usepackage{algorithm}
\usepackage{algorithmic}
\usepackage[normalem]{ulem}
\useunder{\uline}{\ul}{}

\begin{document}

\title{FedGTA: Topology-aware Averaging for Federated Graph Learning}

\author{Xunkai Li}
\affiliation{
\institution{Beijing Institute of Technology}
}
\email{cs.xunkai.li@gmail.com}

\author{Zhengyu Wu}
\affiliation{
\institution{Beijing Institute of Technology}
}
\email{Jeremywzy96@outlook.com}

\author{Wentao Zhang}
\affiliation{
  \institution{Mila - Québec AI Institute \\HEC Montréal}  
}
\email{wentao.zhang@mila.quebec}

\author{Yinlin Zhu}
\affiliation{
\institution{Sun Yat-sen University}
}
\email{ylzhuawesome@163.com}

\author{Rong-Hua Li}
\affiliation{
\institution{Beijing Institute of Technology}
}
\email{lironghuabit@126.com}

\author{Guoren Wang}
\affiliation{
\institution{Beijing Institute of Technology}
}
\email{wanggrbit@126.com}
\begin{abstract}
Federated Graph Learning (FGL) is a distributed machine learning paradigm that enables collaborative training on large-scale subgraphs across multiple local systems. 
Existing FGL studies fall into two categories: 
(i) FGL Optimization, which improves multi-client training in existing machine learning models;
(ii) FGL Model, which enhances performance with complex local models and multi-client interactions. 
However, most FGL optimization strategies are designed specifically for the computer vision domain and ignore graph structure, presenting dissatisfied performance and slow convergence.
Meanwhile, complex local model architectures in FGL Models studies lack scalability for handling large-scale subgraphs and have deployment limitations.
To address these issues, we propose Federated Graph Topology-aware Aggregation (FedGTA), a personalized optimization strategy that optimizes through topology-aware local smoothing confidence and mixed neighbor features. 
During experiments, we deploy FedGTA in 12 multi-scale real-world datasets with the Louvain and Metis split. 
This allows us to evaluate the performance and robustness of FedGTA across a range of scenarios.
Extensive experiments demonstrate that FedGTA achieves state-of-the-art performance while exhibiting high scalability and efficiency.
The experiment includes ogbn-papers100M, the most representative large-scale graph database so that we can verify the applicability of our method to large-scale graph learning.
To the best of our knowledge, our study is the first to bridge large-scale graph learning with FGL using this optimization strategy, contributing to the development of efficient and scalable FGL methods.

\end{abstract}

\maketitle

\pagestyle{\vldbpagestyle}
\begingroup\small\noindent\raggedright\textbf{PVLDB Reference Format:}\\
\vldbauthors. \vldbtitle. PVLDB, \vldbvolume(\vldbissue): \vldbpages, \vldbyear.\\
\href{https://doi.org/\vldbdoi}{doi:\vldbdoi}
\endgroup
\begingroup
\renewcommand\thefootnote{}\footnote{\noindent
This work is licensed under the Creative Commons BY-NC-ND 4.0 International License. Visit \url{https://creativecommons.org/licenses/by-nc-nd/4.0/} to view a copy of this license. For any use beyond those covered by this license, obtain permission by emailing \href{mailto:info@vldb.org}{info@vldb.org}. Copyright is held by the owner/author(s). Publication rights licensed to the VLDB Endowment. \\
\raggedright Proceedings of the VLDB Endowment, Vol. \vldbvolume, No. \vldbissue\ %
ISSN 2150-8097. \\
\href{https://doi.org/\vldbdoi}{doi:\vldbdoi} \\
}\addtocounter{footnote}{-1}\endgroup

\ifdefempty{\vldbavailabilityurl}{}{
\vspace{.3cm}
\begingroup\small\noindent\raggedright\textbf{PVLDB Artifact Availability:}\\
The source code, data, and/or other artifacts have been made available at {\url{https://github.com/xkLi-Allen/FedGTA}}.
\endgroup
}

\section{Introduction}
\label{sec: introdcution}
Graphs are extensively utilized for modeling complex systems, primarily due to their ability to visually represent the relational information between different entities, which sets them apart from other types of data.
Its growing prevalence in recommendation systems~\cite{graphrec1, graphrec2}, drug discovery~\cite{graphdrug1, graphdrug2}, and financial risk control~\cite{graphfinancial1, graphfinancial2} urges the development of a correspondent graph analysis tool. 
Graph Neural Networks (GNNs) emerge as a promising approach to achieve state-of-the-art performance in node-level~\cite{wu2019sgc,Hu2021ahgae,zhang2022gamlp}, edge-level~\cite{Zhang18link_prediction1,cai2021linegnn,besta2022motif}, and graph-level~\cite{xu2018gin,nguyen2022u2gnn,pmlr2022Jacobigcn} downstream tasks.

Researchers in the database community have recently focused on developing centralized data-driven pipelines of large-scale graph learning~\cite{liao2022scara, 10.14778/scara_scalable_gnn_2}. 
However, the surge in retrieved graph data for real-world applications has generated greater interest in the decentralized settings~\cite{nadal2021graphdatabase1, pan2023graphdatabase2, lei2023graphdatabase3} for large-scale graph learning~\cite{hu2020ogb, liu2022federated, khatua2023igb}.
Specifically, collecting data across different locations and sources often requires efforts from different institutions, among which information sharing may be impeded for legal or competitive reasons.
For example, the disease network~\cite{peng2022fedni} and online transaction network~\cite{fu2022federated} require the participation of multiple local clients, such as hospitals and regional institutions.

To address the above issues, one promising solution is Federated Graph Learning (FGL), a distributed training framework for GNNs.
FGL approaches involve devices training their own local models using self-collected data, upon which the central server achieves optimization to obtain a global model.
Despite preserving privacy, FGL can also overcome computational and storage limitations by dealing with large-scale subgraphs through scalable local models and effective optimization strategies. 
In a nutshell, we can summarize the current studies in FGL into the following two types: 
(i) FGL Optimization: applying improved federated optimization methods to the existing graph learning models; 
(ii) FGL Model: designing local model architectures and multi-client interactions.

\begin{figure}[htbp]   
	\centering
    \setlength{\abovecaptionskip}{0.2cm}
    \setlength{\belowcaptionskip}{-0.5cm}
	\includegraphics[width=\linewidth,scale=1.00]{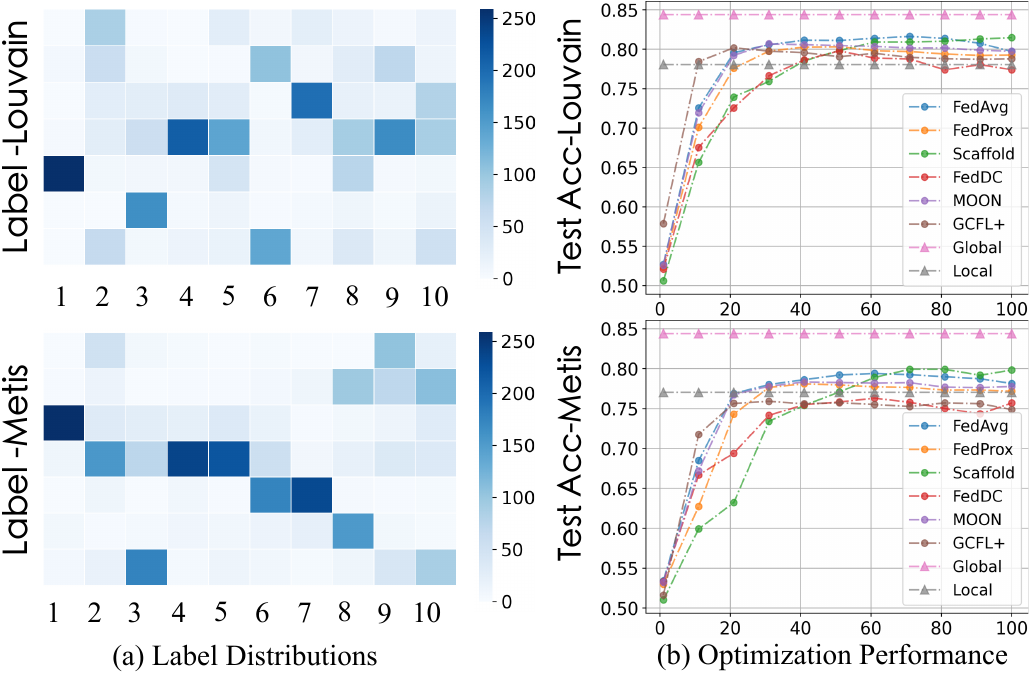}
	\caption{
    Empirical analysis on Cora in 10 clients with GCN, which contains 7 different node labels. 
    (a) The color from white to blue represents the number of nodes held by different clients in each class a gradual increase in quantity.
    (b) The x-axis of the line plot is the federated training round.
    "Global" and "Local" represent the model performance in centralized and siloed settings, respectively.
    }
    \label{fig: empirical analysis}
\end{figure}

A proper optimization strategy is critical in achieving multi-client collaborative training. 
FedAvg~\cite{mcmahan2017fedavg}, a simple yet effective optimization strategy, which performs weighted model aggregation based on the proportional weights of the data size of participating clients against the total combined data size. 
Although FedAvg is topology-independent and primarily intended for CNNs or MLP, many existing FGL Model studies~\cite{zhang2021fedsage,chen2021fedgl,he2021fedgraphnn} still apply it and enhance their performance by well-designed model architectures. 
However, these models have limited scalability, which ultimately restricts their real-world applicability.
To illustrate further, we present an experiment in Fig.~\ref{fig: empirical analysis}, which reveals two major FGL limitations:
(i) lack of investigation of federated subgraph distribution;
(ii) the absence of topology-aware optimization strategies.

The first limitation underscores the label Non-independent identical distribution (Non-iid) problem in FGL as shown in Fig.~\ref{fig: empirical analysis}(a), where we chose Louvain~\cite{blondel2008louvain} and Metis~\cite{karypis1998metis}, two widely applied federated subgraphs simulation methods based on community search. 
As the homogeneity assumption suggests for most real-world graphs, linked nodes are similar in both feature distributions and labels~\cite{graphsurvey1, graphsurvey2, huang2020cs}. 
Such premises lead to the results concluded in Fig.~\ref{fig: empirical analysis}(a). 
We use different colors to highlight the varying numbers of nodes distributed through the data simulation in each client. 
We observe that each client exhibits distinctive or similar label distributions (label Non-iid).
In FGL, the label distributions held by clients are often significantly different, and this property requires extra emphasis since the aggregation between clients regardless of label distribution can lead to unsatisfying results.   
In fact, this is the essential reason for the sub-optimal performance of the existing FGL Model~\cite{cheung2021fedsgc,wu2021fedgnn,chen2021fedgl,zhang2021fedsage,baek2022fedpub}.
To maximize efficiency, we require a personalized optimization strategy that selects clients with similar label distribution instead of applying permutations of all possibilities. 
To the best of our knowledge, it has not yet been specifically addressed.

The second limitation is that most currently adopted optimization strategies overlook the topology of the graph.
In Fig.~\ref{fig: empirical analysis}(b), we conduct a subsequent experiment to detect the label Non-iid problem illustrated in Fig.~\ref{fig: empirical analysis}(a).
The results show that methods like FedProx~\cite{li2020fedprox}, Scaffold~\cite{karimireddy2020scaffold}, MOON~\cite{li2021moon}, and FedDC~\cite{gao2022feddc}, which are applied to solve label Non-iid problems in the computer vision domain, do not achieve competitive results compared to FedAvg and local train.
GCFL+~\cite{xie2021gcfl} assumes that the graph topology is implicitly incorporated into the local model through uploaded gradients, but this approach fails to fully capture the topology and leads to sub-optimal performance.
These results motivate us to investigate why conventional methods that perform well in the computer vision domain fail to replicate their success in FGL. 
Our conclusion is that these methods do not directly consider the topology properties, which are crucial elements in graph studies.
In light of this finding, our main motivation is to design a topology-aware optimization strategy that specifically targets the label Non-iid problem in the federated graph collaborative training process.

In this paper, we propose Federated Graph Topology-aware Aggregation (FedGTA), a novel and scalable optimization strategy, for FGL and present state-of-the-art performance in efficiency.
Specifically, each client calculates topology-aware local smoothing confidence and mixed moments of neighbor features and uploads them with correspondent model weights to the server. 
Then, the server is able to customize the optimization strategy for each participating client to continue federated training. 

In summary, the main contributions of this paper are:
(1) \textbf{Problem Connection.} We introduce a novel perspective for integrating large-scale graph learning with FGL.
(2) \textbf{New Method.} We propose FedGTA, a novel topology-aware optimization strategy for FGL. 
It has been formulated into a unified framework that can be applied to any graph learning model. 
(3) \textbf{SOTA Performance.} We conduct experiments on 12 real-world benchmark datasets including the archetypal large-scale graph dataset, ogbn-papers100M, with prevalent GNNs. 
We demonstrate that FedGTA significantly outperforms the state-of-the-art baselines on both performance and efficiency.

\begin{table*}[htbp]
\caption{Algorithm analysis for existing FGL Optimization/Model studies. 
$n$, $m$, $c$, and $f$ are the number of nodes, edges, classes, and feature dimensions, respectively. 
$s$ is the number of selected augmented nodes and $g$ is the number of generated neighbors.
$b$ and $T$ are the batch size and dynamic training round, respectively.
$k$ and $K$ correspond to the number of times we aggregate features and moments order, respectively. 
Besides, $N$ is the number of participating clients in each training round.
For model-agnostic optimization strategies, we choose SGC as the local model ($k$-step feature propagation), and for FGL methods, we adopt the model architecture ($L$-layer) used in their original paper.
}
\footnotesize 
\label{tab: algorithm_analysis}
\resizebox{\textwidth}{18mm}{
\setlength{\tabcolsep}{1.2mm}{

\begin{tabular}{c|c|ccc|ccc}
\midrule[0.3pt]
Method        & Type   & Client Mem.         & Server Mem.       & Inference Mem.      & Client Time.            & Server Time.            & Inference Time      \\ \midrule[0.3pt]
FedAvg        & Optim. & $O((b+k)f+f^2)$     & $O(N+Nf^2)$       & $O((b+k)f+f^2)$     & $O(kmf+nf^2)$           & $O(N)$                  & $O(kmf+nf^2)$       \\
FedProx       & Optim. & $O((b+k)f+2f^2)$    & $O(N+Nf^2)$       & $O((b+k)f+f^2)$     & $O(kmf+nf^2+f^2)$       & $O(N)$                  & $O(kmf+nf^2)$       \\
Scaffold      & Optim. & $O((b+k)f+2f^2)$    & $O(N+2Nf^2+f^2)$  & $O((b+k)f+f^2)$     & $O(kmf+nf^2+f^2)$       & $O(N+Nf^2+f^2)$         & $O(kmf+nf^2)$       \\
MOON          & Optim. & $O((3b+k)f+f^2)$    & $O(N+Nf^2)$       & $O((b+k)f+f^2)$     & $O(kmf+nf^2+2nf)$       & $O(N)$                  & $O(kmf+nf^2)$       \\
FedDC         & Optim. & $O((b+k)f+4f^2)$    & $O(N+2Nf^2)$      & $O((b+k)f+f^2)$     & $O(kmf+nf^2+4f^2)$      & $O(N)$                  & $O(kmf+nf^2)$       \\
GCFL+         & Optim. & $O((b+k)f+f^2)$     & $O(N+Nf^2+TNf^2)$ & $O((b+k)f+f^2)$     & $O(kmf+nf^2)$           & $O(N+N^2(\log(N)+T^2f^2))$ & $O(kmf+nf^2)$       \\ \midrule[0.3pt]
FedGL         & Model  & $O(Lnf+Lf^2+n^2)$   & $O(N+NLf^2)$      & $O(Lnf+Lf^2+n^2)$   & $O(Lmf+Lnf^2+n^2f)$     & $O(N)$                  & $O(Lmf+Lnf^2+n^2f)$ \\
FedSage+      & Model  & $O(L(n+sg)f+3Lf^2)$ & $O(N+3NLf^2)$     & $O(L(n+sg)f+3Lf^2)$ & $O(L(m+sg)f+L(n+sg)f^2)$     & $O(N)$                  & $O(L(m+sg)f+L(n+sg)f^2)$ \\ \midrule[0.3pt]
FedGTA (ours) & Optim. & $O((b+k)f+f^2+kKc)$ & $O(N+Nf^2+NkKc)$  & $O((b+k)f+f^2)$     & $O(km(f+knc)+n(f^2+c))$& $O(N+NkKc)$             & $O(kmf+nf^2)$       \\ \bottomrule[0.3pt]
\end{tabular}
}}

\end{table*}

\section{PRELIMINARIES AND RELATED WORKS}
In this section, we first describe the notations and problem formulation in this paper.
Then we briefly discuss the difference between conventional and scalable GNNs and the FGL Optimization/Model studies.
Meanwhile, summarized in Table~\ref{tab: algorithm_analysis}, we present an analysis on the complexity bounds of existing FGL studies.

\subsection{Problem Formulation}
Consider a graph $G = (\mathcal{V}, \mathcal{E})$ with $|\mathcal{V}|=n$ nodes and $|\mathcal{E}|=m$ edges, the adjacency matrix (including self-loops) is $\hat{\mathbf{A}}\in\mathbb{R}^{n\times n}$, the feature matrix is $\mathbf{X} = \{x_1,\dots,x_n\}$ in which $x_v\in\mathbb{R}^{f}$ represents the feature vector of node $v$, and $f$ represents the dimension of the node attributes.
Besides, $\mathbf{Y} = \{y_1,\dots,y_n\}$ is the label matrix, where $y_v\in\mathbb{R}^{|\mathcal{Y}|}$ is a one-hot vector and $|\mathcal{Y}|$ represents the number of the classes.
The semi-supervised node classification task is based on the topology of labeled set $\mathcal{V}_L$ and unlabeled set $\mathcal{V}_U$, and the nodes in $\mathcal{V}_U$ are predicted with the model supervised by $\mathcal{V}_L$.

\subsection{Conventional and Scalable GNNs}
\label{sec:simple_scalable_gnns}
Graph Neural Networks (GNNs) adopt spectral graph theory and deep learning to enable graph learning. 
Specifically, propagation operators are defined based on topology, while trainable weights are applied to learn node attributes. 
In this study, we present two standards of GNNs that differ in model architecture design and their corresponding implications in real-world application scenarios.

\underline{\emph{Conventional GNNs}}.
GCN~\cite{welling2016gcn} and GAT~\cite{velivckovic2017gat} are two widely used methods that employ coupled message-passing schemes to propagate information across the nodes. 
However, when dealing with real-world applications involving large-scale graphs, scalability becomes a major concern due to their limited capacity. 

\underline{\emph{Scalable GNNs}}.
There are two major approaches to achieving GNN scalability. 
One is developing sampling-based GNNs, such as 
GraphSAGE~\cite{hamilton2017graphsage} randomly samples neighbors for computation in each mini-batch, Fast-GCN~\cite{chen2018fastgcn} samples a fixed number of nodes at each layer, and Cluster-GCN~\cite{chiang2019clustergcn} is implemented based on graph-level clustering. 
However, recent studies~\cite{feng2022grand+,zhang2022pasca,zhang2022gamlp} emphasize a decoupled mechanism due to its simple operative mechanism and superior performance.
For example, SGC~\cite{wu2019sgc} reduces GNNs into a linear model operating on $k$-layers propagated features $\mathbf{X}^{(k)}$:
    \begin{equation}
        \label{eq:sgc}
        \mathbf{X}^{(k)}=\tilde{\mathbf{A}}^k \mathbf{X}^{(0)}, \;\tilde{\mathbf{A}} = \hat{\mathbf{D}}^{r-1}\hat{\mathbf{A}}\hat{\mathbf{D}}^{-r},\;\mathbf{Y}=\operatorname{softmax}\left(\boldsymbol{\Theta} \mathbf{X}^{(k)}\right),
    \end{equation}
where $\mathbf{X}^{(0)}=\mathbf{X}$, $\hat{\mathbf{D}}$ is the degree matrix of $\hat{\mathbf{A}}$, $r\in[0,1]$ denotes the propagation kernel coefficient, and $\mathbf{W}$ represents weight matrix.
By default $r=0.5$, we can get the symmetric normalization adjacency matrix $\hat{\mathbf{D}}^{-1/2}\hat{\mathbf{A}}\hat{\mathbf{D}}^{-1/2}$~\cite{gasteiger2018ppnp}.
As the propagated features $\mathbf{X}^{(k)}$ can be precomputed, SGC is easy to scale to large graphs.
Inspired by it, SIGN~\cite{frasca2020sign} proposes to concatenate the learnable propagated features: $\left[\mathbf{X}^{(0)} \mathbf{W}_0, \ldots, \mathbf{X}^{(k)} \mathbf{W}_k\right]$.
S$^2$GC~\cite{zhu2021ssgc} proposes to average the spectral features: $\mathbf{X}^{(k)}=\sum_{l=0}^k \tilde{\mathbf{A}}^l \mathbf{X}^{(0)}$.
GBP~\cite{chen2020gbp} further utilizes the $\beta$ weighted averaging: $\mathbf{X}^{(k)}=\sum_{l=0}^k w_l \tilde{\mathbf{A}}^l \mathbf{X}^{(0)}, w_l=\beta(1-\beta)^l$.
GAMLP~\cite{zhang2022gamlp} achieves information aggregation based on the attention mechanisms ${\mathbf{X}}^{(k)}=\tilde{\mathbf{A}}^k\mathbf{X}^{(0)} \| \sum_{l=0}^{k-1} w_l \mathbf{X}^{(l)}$, where attention weight $w_l$ has multiple calculation versions.

\subsection{FGL Optimization}
\label{sec:optimization}
FedAvg~\cite{mcmahan2017fedavg} is the most widely used optimization strategy for FL. 
It implements model aggregation by using a simple weighted average of the model parameters received from each participating client. 
The weights are proportional to the training data size. 
Its generic form with $N$ participating clients and learning rate $\eta$ is defined as
\begin{equation}
    \small
    \label{eq:localtrain}
    \begin{aligned}
    \widetilde{\mathbf{W}}^{t} &= \sum_{i=1}^N\frac{{n}_i}{{n}}\mathbf{W}^{t-1}_i,\forall i, \mathbf{W}^{t-1}_{i}= \widetilde{\mathbf{W}}^{t-1} - \eta\nabla f,\\
    \mathbf{W}_i^{t}&=\widetilde{\mathbf{W}}^{t}-\eta\nabla f\left(\widetilde{\mathbf{W}}^t,(\mathbf{A}_i,\mathbf{X}_i,\mathbf{Y}_i)\right)\\
    &=\widetilde{\mathbf{W}}^{t}+\eta\sum_{i \in \mathcal{V}_l} \sum_j \mathbf{Y}_{i j} \log \left(\operatorname{softmax}(\hat{\mathbf{Y}})_{i j}\right),
    \end{aligned}
\end{equation}
where ${n}_i$ and ${n}$ represent the $i$-th client and the global data size, $\nabla f(\cdot)$ represents the gradients.
It can be obtained by any reasonable loss function that evolves with downstream tasks.
$\mathbf{W}_i^t$ and $\widetilde{\mathbf{W}}^t$ represent the local model weights held by the $i$-th client in round $t$ and the aggregated model weights received from the server.
 
Despite its effectiveness, FedAvg fails to solve the weight-shifting problem caused by Non-iid data.
To address this challenge, several methods have been proposed, but they largely focus on the computer vision domain rather than on the graphs.
FedProx~\cite{li2020fedprox} limits the deviation of the local model from the global model.
Scaffold~\cite{karimireddy2020scaffold} introduces server and client control variables to control the model's updated direction. 
MOON~\cite{li2021moon} introduces model-contrastive loss in the local training.
FedDC~\cite{gao2022feddc} utilizes learnable local drift variables to bridge the above gap.
GCFL+~\cite{xie2021gcfl} utilizes weight clustering techniques to custom model aggregation in graph classification.

\subsection{FGL Model}
Graph data has been demonstrated to be superior in multiple applications. 
Recently, FGL~\cite{liu2022fglsurvey, wang2022fsg, he2021fedgraphnn} has received a lot of attention due to its unique advantage in training GNN models collaboratively without sharing collected data for safety concerns. 

As mentioned in Section~\ref{sec: introdcution}, there are two main strategies for improving FGL studies. 
The first strategy involves optimizing the FGL process and applying it to existing GNNs. 
We discussed some previous works related to this in Section~\ref{sec:optimization}. 
The second strategy involves improving the model architectures and using FedAvg as the default method, presented by FedSage, FedGNN, and FedGL.
FedSage+~\cite{zhang2021fedsage} implements local subgraph augmentation via the missing neighbor generator.
FedGNN~\cite{wu2021fedgnn} attempts to propose a federated graph recommendation model with security guarantees. 
FedGL~\cite{chen2021fedgl} proposes to use the overlapping subgraph nodes to implement global supervision among multi-clients. 

To further illustrate, we provide the algorithmic complexity of each method in Table~\ref{tab: algorithm_analysis}, where "Optim." represents FGL Optimization and "Model" denotes FGL Models. 
For the $k$-layer SGC model with batch size $b$, the precomputed results are bounded by a space complexity of $O((b+k)f)$. 
The overhead for linear regression is $O(f^2)$.
For the server performing FedAvg, it needs to receive the model weights and the number of samples participating in this round.
Its space complexity and time complexity are bounded as $O(N+Nf^2)$ and $O(N)$.
As discovered by previous studies~\cite{chen2020gbp,zhang2022gamlp,zhang2022pasca}, the dominating term is $O(kmf)$ or $O(Lmf)$ when the graph is large since feature learning can be accelerated by parallel computation. 
The full large-graph propagation becomes extremely difficult and thus, FedGL and FedSage+ lead to unacceptable space-time overhead because of the $O(n^2)$ term and $O(m+n+2sg)$ term.

\begin{figure*}[htbp]   
	\centering
    \setlength{\abovecaptionskip}{0.4cm}
	\includegraphics[width=\linewidth,scale=1.00]{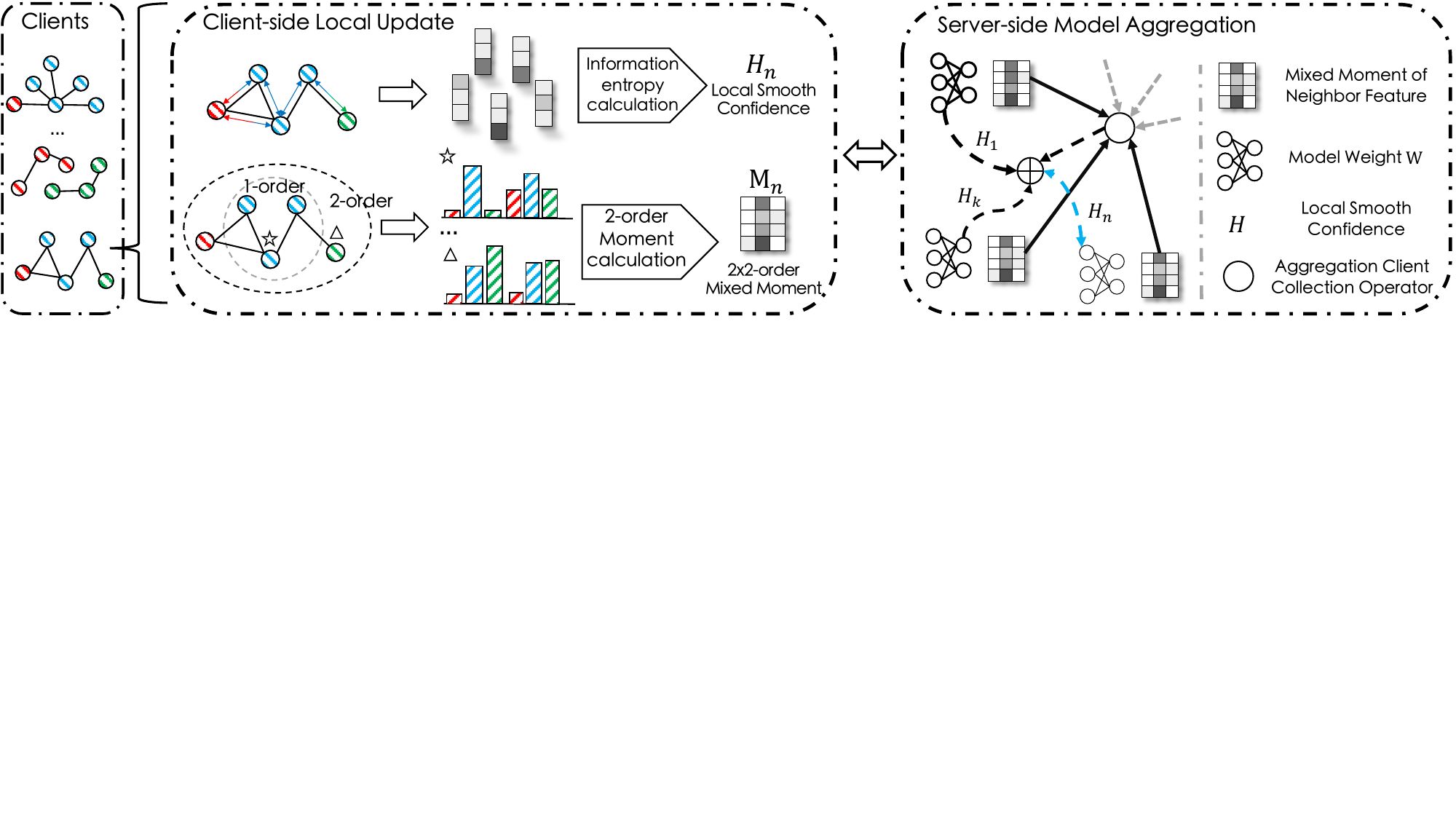}
 
	\caption{
    Overview of our proposed FedGTA framework.
    The different colors of the nodes represent the different labels.}
	\label{fig:framework}
\end{figure*}

\section{FedGTA FRAMEWORK}
As an optimization strategy, FedGTA provides a novel perspective that integrates both large-scale graph learning and FGL into its design as shown in Fig.~\ref{fig:framework}.
To begin with, on the \textbf{Client-side}, participating clients encode topology and node attributes. 
After that, they calculate the local smoothing confidence and mixed moments of neighbor features, which are then uploaded to the server. 
On the \textbf{Server-side}, FedGTA performs personalized model aggregation for each client based on the mixed moments of neighbor features and uses the local smoothing confidence as the aggregation weights.

\emph{\underline{Local Smoothing Confidence.}}
Since GNNs trained by the smoothing graph are more confident with their prediction~\cite{gasteiger2018ppnp,huang2020cs,zhang2022pasca}, we grant it with higher contribution in the model aggregation process. 

\emph{\underline{Mixed Moments of Neighbor Features.}} 
By using the mixed moments of neighbor features to measure the distribution of subgraphs, we can limit the model aggregation process to only those clients with similar subgraph distributions. 

\subsection{The Proposed FedGTA}
To encode topology and node attributes, we introduce $k$-step Non-parameters Label Propagation (Non-param LP), which establishes relationships among the current node and its $k$-hop neighbors
\begin{equation}
\label{eq:nonparamlp}
    \begin{aligned}
    \hat{\mathbf{Y}}&=\mathrm{Softmax}\left(\mathrm{Encoder(\mathbf{A,X})}\right),\\
    \hat{\mathbf{Y}}^k(v_i)&=\alpha\hat{\mathbf{Y}}^0(v_i)+(1-\alpha)\sum_{j\in\mathcal{N}_i}\frac{1}{\sqrt{\tilde{d}_i\tilde{d}_j}}\hat{\mathbf{Y}}^{k-1}(v_j),
    \end{aligned}
\end{equation}
where $\mathrm{Encoder}(\cdot,\cdot)$ represent any embedding model and $\hat{\mathbf{Y}}\in\mathbb{R}^{n\times|\mathcal{Y}|}$ denotes the soft label matrix.
We follow the approximate calculation of the personalized PageRank~\cite{gasteiger2018ppnp}, where $\mathcal{N}_i$ represents the one-hop neighbors of $i$.
We default set $\alpha = 1/2, k=5$ to encode deep structural information.
Then, we obtain the topology-aware soft label matrix.
Based on this, each client calculates the quantitative metrics to perform model optimization on the server side.
The complete algorithm can be obtained as described in Algorithm\ref{alg:clientupdate}.


\emph{\underline{Client-Local Smoothing Confidence.}}
The key insight is to quantify smoothness by the entropy of local predictions
\begin{equation}
\label{eq:localsmoothconfidence}
    \begin{aligned}
    H=\sum_{i=1}^{|\mathcal{V}|}\sum_{j=1}^{|\mathcal{Y}|}\mathbf{D}_{ii}\left(e^{-1}-\left(-\hat{\mathbf{Y}}^k_{{ij}}\log\hat{\mathbf{Y}}^k_{{ij}}\right)\right).
    \end{aligned}
\end{equation}

Since the model based on the smoothing subgraph tends to produce clearer predictions with a lower entropy value, modification is needed as we want to represent higher confidence figuratively. 
Therefore we subtract it from the theoretical maximum $e^{-1}$, and then we obtain $H$ by considering both the local neighbors and the number of samples, with the addition of the degree matrix $\mathbf{D}_{ii}$.

\begin{algorithm}[t]
\caption{FedGTA-Client Update} 
\label{alg:clientupdate}
\begin{algorithmic}[1] 
    \FOR {each communication round $t = 1, ..., T$}
    \FOR {each local model update $e = 1, ..., E$}
    \STATE Update local model weights $\mathbf{W}$ according to the Eq~(\ref{eq:localtrain});
    \ENDFOR
    \STATE Calculate the topology-aware label distribution by Eq~(\ref{eq:nonparamlp});
    \STATE \underline{\emph{Local\;Smoothing\;Confidence}}:
    \STATE Calculate the entropy of soft label predictions;
    \STATE Execute smoothness quantification $H$ based on the Eq~(\ref{eq:localsmoothconfidence});
    \STATE \underline{\emph{Mixed\;Moments\;of\;Neighbor\;Features}}:
    \STATE Calculate the $K$-order moments $\mathbf{M}$ based on the Eq~(\ref{eq:mixedmoments});
    \STATE Each client uploads the relevant $H$, $\mathbf{M}$, and model weight $W$;
    \ENDFOR
\end{algorithmic}
\end{algorithm}

\begin{algorithm}[t]
\caption{FedGTA-Server Aggregation} 
\label{alg:serveraggregation}
\begin{algorithmic}[1] 
    \FOR {each communication round $t = 1, ..., T$}
    \FOR{each client $i=1,\dots,N$}
    \STATE Calculate the set of model aggregation for the current client $i$ based on $\mathbf{M}$, $\epsilon$, and Eq~(\ref{eq:sim});
    \STATE Execute federated model optimization based on the model aggregation sets $\mathbb{I}_i$, $H_i$, and Eq~(\ref{eq:fedgta}) to get $\widetilde{\mathbf{W}}_i$;
    \STATE Server sends global model $\widetilde{\mathbf{W}}_i$ to each local client $i$;
    \ENDFOR
    \ENDFOR
\end{algorithmic}
\end{algorithm}

\emph{\underline{Client-Mixed Moments of Neighbor Features.}}
Here we introduce the mixed moments of neighbor features, which is used to generalize local subgraph.
Specifically, we compute the $K$-order mixed moments of $k$-step propagated soft labels $\mathbf{M}\in\mathbb{R}^{(k\times K)\times|\mathcal{Y}|}$.
We present the formal representation of central moments as an example
\begin{equation}
\label{eq:mixedmoments}
    \begin{aligned}
    \mathbf{M}(\hat{y}^k_i)&=\mathbb{E}\left(\left(\hat{y}_i^{1}-\mu_i^{1}\right)^1\right)||\dots||\mathbb{E}\left(\left(\hat{y}_i^{k}-\mu_i^{k}\right)^K\right),\\
    \mathbb{E}\left(\left(\hat{y}_i^{k}-\mu_i^{k}\right)^K\right)&=\left(\alpha\hat{y}_i^0+(1-\alpha)\sum_{j\in\mathcal{N}_i}\frac{1}{\sqrt{\tilde{d}_i\tilde{d}_j}}\hat{y}_j^{k-1}-\frac{1}{|\mathcal{Y}|}\sum\hat{y}_i^{k}\right)^K,
    \end{aligned}
\end{equation}
where $\cdot||\cdot$ is concatenation and $\mu_i^1$ denotes the mean value of $\hat{y}_i^{1}$.
Due to the differences in the subgraphs held by clients, it becomes imperative to employ appropriate quantification metrics, guiding the server toward performing model aggregation tailored to each client. 
Recognizing the significance of topology in graphs and its potential correlation with features, we utilize the mixed moments of neighbor features to realize topology-aware quantification of discrepancies among multi-client subgraphs.
After that, participating clients upload their metrics and model weights to the server.

\emph{\underline{Server-Model Aggregation.}}
Based on the above metrics, the server computes the similarity of the mixed moments of neighbor features and utilizes a data-driven context threshold to obtain the set of other clients for which the current client performs model aggregation.
The intuition is to avoid clients with high variability interfering with each other during the federated collaborative training process.
The computation process is formally defined as
\begin{equation}
\label{eq:sim}
    \begin{aligned}
    &\mathbb{I}_i=\{j|\mathrm{sim}(i,j)\ge \epsilon\}\cup i,\;\forall i,j\in \mathrm{Set}(N),\;j\neq i,\\
    &\mathrm{sim}(i,j)=\frac{\sum_{p=1}^{k\times K}\mathbf{M}_i^p\cdot\mathbf{M}_j^p}{\sqrt{\sum_{p=1}^{k\times K}(\mathbf{M}_i^p)^2}\sqrt{\sum_{p=1}^{k\times K}(\mathbf{M}_j^p)^2}},\\
    \end{aligned}
\end{equation}
where $\epsilon$ denotes the threshold and $\mathbb{I}_i$ represents the model aggregation set for client $i$.
$\mathrm{Set}(N)$ represents the set of all participating clients in the current round.
In fact, the cosine similarity in Eq~(\ref{eq:sim}) can be replaced with any reasonable metric.
Based on this, the server performs personalized weighted model aggregation for each participating client based on the local smoothing confidence
\begin{equation}
\label{eq:fedgta}
    \begin{aligned}
    &\forall i,\widetilde{\mathbf{W}}_i^{t+1} \leftarrow \sum_{j,k\in\mathbb{I}_i}\frac{H_i}{\sum_{} H_j}\mathbf{W}^{t}_j,\; \mathbf{W}^{t}_{j} \leftarrow \widetilde{\mathbf{W}}^{t}_{j} - \eta\nabla f.
    \end{aligned}
\end{equation}
Notably, with the involvement of mixed moments of neighbor features, the model aggregation process based on the comparison of local smoothing confidence can become more accurate and efficient as it will only consider the client with similar subgraph distribution. 
The complete algorithm can be referred to as Algorithm\ref{alg:serveraggregation}.

\subsection{Complexity Analysis}

We provide the complexity analysis of our proposed FedGTA and other FGL Optimization/Model studies in Table~\ref{tab: algorithm_analysis}.
For the client side of FedGTA, the computation complexity for calculating the entropy for each row of soft label $\hat{\mathbf{Y}}\in\mathbb{R}^{n\times c}$ is $O(nc)$, and the computation complexity for moments is $O(k^2mnc)$. 
On the server side, the computation complexity for calculating the cosine similarity of $N$ clients' moments is $O(NkKc)$.
The computation complexity of our method only depends on the training-independent model-agnostic sparse matrix multiplication, while other FGL optimization strategies adopt coupled training mechanisms, whose additional loss terms lead to excessive computation cost accompanied by the local training process.
More analysis and experiments of algorithmic complexity can be referred to Section~\ref{sec:Efficiency and Scalability Analysis}.

\begin{table*}[htbp]
\setlength{\abovecaptionskip}{0.2cm}
\caption{The statistical information of the experimental datasets.
}
\footnotesize 
\label{tab: datasets}
\resizebox{\linewidth}{27.5mm}{
\setlength{\tabcolsep}{2mm}{
\begin{tabular}{cccccccc}
\midrule[0.3pt]
Dataset          & \#Nodes     & \#Features & \#Edges       & \#Classes & \#Train/Val/Test  & \#Task      & Description         \\ \midrule[0.3pt]
Cora             & 2,708       & 1,433      & 5,429         & 7         & 20\%/40\%/40\%    & Transductive & citation network    \\
CiteSeer         & 3,327       & 3,703      & 4,732         & 6         & 20\%/40\%/40\%    & Transductive & citation network    \\
PubMed           & 19,717      & 500        & 44,338        & 3         & 20\%/40\%/40\%    & Transductive & citation network    \\ \midrule[0.3pt]
Amazon Photo     & 7,487       & 745        & 119,043       & 8         & 20\%/40\%/40\%    & Transductive & co-purchase graph   \\
Amazon Computer  & 13,381      & 767        & 245,778       & 10        & 20\%/40\%/40\%    & Transductive & co-purchase graph   \\
Coauthor CS      & 18,333      & 6,805      & 81,894        & 15        & 20\%/40\%/40\%    & Transductive & co-authorship graph \\
Coauthor Physics & 34,493      & 8,415      & 247,962       & 5         & 20\%/40\%/40\%    & Transductive & co-authorship graph \\ \midrule[0.3pt]
ogbn-arxiv       & 169,343     & 128        & 2,315,598     & 40        & 60\%/20\%/20\%    & Transductive & citation network    \\
ogbn-products    & 2,449,029   & 100        & 61,859,140    & 47        & 10\%/5\%/85\%     & Transductive & co-purchase graph   \\
ogbn-papers100M  & 111,059,956 & 128        & 1,615,685,872 & 172       & 1200k/200k/146k   & Transductive & citation network    \\ \midrule[0.3pt]
Flickr           & 89,250      & 500        & 899,756       & 7         & 44k/22k/22k       & Inductive    & image network       \\
Reddit           & 232,965     & 602        & 11,606,919    & 41        & 155k/23k/54k      & Inductive    & social network      \\ \midrule[0.3pt]
\end{tabular}
}}
\end{table*}

\begin{table*}[]
\setlength{\abovecaptionskip}{0.3cm}
\caption{Transductive performance on FGL Optimization/Model studies.
"OOM" stands for out-of-memory error.
"Global" represents the training and inference using the complete global graph under centralized conditions.
The best result is \textbf{bold}.
The second result is \ul{underlined}.
}
\footnotesize 
\label{tab: trans_cmp}
\resizebox{\linewidth}{36.5mm}{
\setlength{\tabcolsep}{1.4mm}{

\begin{tabular}{cc|cccccccccc}
\midrule[0.3pt]
Model                  & Optimization & Cora              & CiteSeer          & PubMed            & \begin{tabular}[c]{@{}c@{}}Amazon\\ Photo\end{tabular} & \begin{tabular}[c]{@{}c@{}}Amazon\\ Computer\end{tabular} & \begin{tabular}[c]{@{}c@{}}Coauthor\\ CS\end{tabular} & \begin{tabular}[c]{@{}c@{}}Coauthor\\ Physics\end{tabular} & \begin{tabular}[c]{@{}c@{}}ogbn\\ arxiv\end{tabular} & \begin{tabular}[c]{@{}c@{}}ogbn\\ products\end{tabular} & \begin{tabular}[c]{@{}c@{}}ogbn\\ papers100M\end{tabular} \\ \midrule[0.3pt]
\multirow{8}{*}{GCN}   & Global       & 84.6±0.3          & 72.1±0.2          & 90.3±0.1          & 92.8±0.3                                               & 84.3±0.4                                                  & 92.5±0.2                                              & 93.1±0.6                                                   & 73.8±0.3                                             & 76.3±0.2                                                & OOM                                                       \\
                       & FedAvg       & 80.7±0.3          & 68.4±0.3          & 85.9±0.1          & 89.6±0.5                                               & 80.3±0.4                                                  & 87.4±0.3                                              & 88.5±0.5                                                   & 66.7±0.4                                             & 71.7±0.2                                                & 58.4±0.3                                                  \\
                       & FedProx      & 80.5±0.2          & 68.7±0.3          & 85.8±0.1          & 88.8±0.7                                               & 80.5±0.6                                                  & 87.5±0.6                                              & 88.6±0.7                                                   & 67.1±0.5                                             & {\ul 72.6±0.3}                                          & 58.7±0.4                                                  \\
                       & Scaffold     & {\ul 81.3±0.4}    & 68.3±0.3          & 85.7±0.2          & 89.5±0.8                                               & 80.4±0.5                                                  & 86.1±0.5                                              & {\ul 89.3±0.7}                                             & 66.9±0.7                                             & 72.4±0.3                                                & 58.8±0.5                                                  \\
                       & MOON         & 80.7±0.4          & {\ul 69.0±0.3}    & 85.8±0.1          & 89.3±0.8                                               & 80.2±0.5                                                  & 87.7±0.5                                              & 89.2±0.7                                                   & 67.3±0.5                                             & 72.3±0.2                                                & 58.2±0.5                                                  \\
                       & FedDC        & 81.0±0.2          & 68.4±0.2          & {\ul 86.2±0.2}    & 89.7±0.6                                               & {\ul 80.8±0.5}                                            & {\ul 87.8±0.5}                                        & 89.1±0.7                                                   & {\ul 67.5±0.5}                                       & 72.0±0.2                                                & {\ul 58.8±0.6}                                            \\
                       & GCFL+        & 80.5±0.1          & 68.1±0.2          & 85.0±0.1          & {\ul 89.9±0.4}                                         & 79.4±0.3                                                  & 87.4±0.2                                              & 88.6±0.3                                                   & 66.8±0.2                                             & 71.7±0.2                                                & 58.0±0.2                                                  \\
                       & FedGTA       & \textbf{82.1±0.3} & \textbf{70.6±0.3} & \textbf{88.0±0.1} & \textbf{91.4±0.7}                                      & \textbf{82.7±0.5}                                         & \textbf{90.0±0.2}                                     & \textbf{91.2±0.5}                                          & \textbf{70.3±0.4}                                    & \textbf{74.8±0.3}                                       & \textbf{60.6±0.3}                                         \\ \midrule[0.3pt]
\multirow{8}{*}{GAMLP} & Global       & 85.7±0.5          & 75.9±0.4          & 91.1±0.1          & 93.1±0.5                                               & 86.0±0.6                                                  & 93.7±0.4                                              & 93.6±1.0                                                   & 80.5±0.6                                             & 84.2±0.3                                                & 68.8±0.1                                                  \\
                       & FedAvg       & 82.2±0.3          & 70.7±0.4          & 86.9±0.1          & 90.4±0.4                                               & 80.6±0.3                                                  & 89.3±0.4                                              & 89.2±0.5                                                   & 71.4±0.7                                             & 79.0±0.3                                                & 62.1±0.2                                                  \\
                       & FedProx      & 82.0±0.8          & 70.6±0.6          & 86.8±0.1          & 90.3±0.6                                               & 80.7±0.5                                                  & 88.7±0.5                                              & 89.3±1.0                                                   & 72.3±1.1                                             & 78.8±0.5                                                & {\ul 63.2±0.3}                                            \\
                       & Scaffold     & 82.6±0.7          & 71.1±0.5          & 86.5±0.2          & 89.8±0.8                                               & {\ul 80.8±0.8}                                            & 88.6±0.6                                              & 89.4±0.9                                                   & 71.6±0.8                                             & 79.2±0.5                                                & 63.1±0.4                                                  \\
                       & MOON         & 81.9±0.4          & 70.9±0.4          & {\ul 87.2±0.2}    & {\ul 90.5±0.7}                                         & 80.5±0.7                                                  & 89.2±0.5                                              & {\ul 90.0±0.8}                                             & {\ul 72.5±0.9}                                       & {\ul 79.3±0.5}                                          & 62.7±0.3                                                  \\
                       & FedDC        & {\ul 83.0±0.5}    & 70.8±0.5          & 87.0±0.1          & 89.6±0.7                                               & 80.8±0.6                                                  & 88.8±0.6                                              & 89.8±1.0                                                   & 71.9±0.8                                             & 78.9±0.5                                                & 63.0±0.3                                                  \\
                       & GCFL+        & 82.7±0.7          & {\ul 71.6±0.3}    & 86.5±0.1          & 90.5±0.4                                               & 80.4±0.3                                                  & {\ul 89.5±0.3}                                        & 89.2±0.4                                                   & 71.5±0.3                                             & 78.5±0.2                                                & 63.0±0.2                                                  \\
                       & FedGTA       & \textbf{83.8±0.6} & \textbf{74.3±0.6} & \textbf{88.4±0.1} & \textbf{91.5±0.5}                                      & \textbf{83.9±0.4}                                         & \textbf{91.2±0.4}                                     & \textbf{91.8±0.5}                                          & \textbf{74.3±0.7}                                    & \textbf{81.6±0.4}                                       & \textbf{66.5±0.3}                                         \\ \midrule[0.3pt]
FedGL                  & FedAvg       & 81.1±0.6          & 70.6±0.6          & 86.5±0.4          & 89.7±1.0                                               & 81.7±0.8                                                  & 88.4±0.8                                              & 88.8±1.2                                                   & 71.4±1.5                                             & OOM                                                     & OOM                                                       \\
FedSage+               & FedAvg       & 82.7±0.9          & 72.0±1.0          & 87.1±0.5          & 90.7±1.3                                               & 82.4±1.5                                                  & 89.2±1.4                                              & 90.0±1.6                                                   & 71.1±1.8                                             & OOM                                                     & OOM                                                       \\ \midrule[0.3pt]
\end{tabular}
}}
\end{table*}

\section{Experiments}
In this section, we conduct a wide range of experiments to verify the effectiveness of FedGTA.
To begin with, we introduce 12 graph benchmark datasets as the global graph and two subgraph simulation strategies widely used in the FGL.
Then, we introduce the baseline backbone GNNs and FGL approaches and detailed experimental setup.
After that, we aim to answer the following questions:
\textbf{Q1}: Compared with other state-of-the-art FGL Optimization/Model studies, can FedGTA achieve better performance?
\textbf{Q2}: What is the generalization ability of FedGTA in the field of FGL?
\textbf{Q3}: Where does the performance gain of FedGTA come from?
\textbf{Q4}: How does FedGTA perform in terms of efficiency and scalability?

\subsection{Experimental Setup}
\textbf{Datasets.}
For a comprehensive comparison, we evaluate FedGTA and other baselines under both transductive and inductive settings. 
For transductive settings, we conduct experiments on 3 small-scale citation networks (Cora, Citeseer, PubMed)~\cite{yang2016cora}, 2 medium-scale user-item datasets (Amazon Computer, Amazon Photo), 2 medium-scale Coauthor datasets (Coauthor CS, Coauthor Physics)~\cite{shchur2018amazon_datasets}, and 3 large-scale OGB datasets (ogbn-arxiv, ogbn-products, ogbn-papers100M)~\cite{hu2020ogb}. 
For inductive settings, we conduct experiments on 2 datasets of medium and large scales: Flickr and Reddit~\cite{zeng2019graphsaint}. 
More details about the above datasets can be found in Table~\ref{tab: datasets}.
Based on this, we provide Louvain~\cite{blondel2008louvain} and Metis~\cite{karypis1998metis} split, which are widely used in FGL~\cite{he2021fedgraphnn,wang2022fsg,zhang2021fedsage,baek2022fedpub}.
Specifically, we apply Louvain on the global graph to assign discovered communities to multi-clients. 
In the Metis split, we assign nodes to each client based on the given number of clients. 
Notably, since the ogbn-papers100M dataset contains a large number of unlabeled nodes, we only perform Louvain split on it. 
This is because we can control the amount of labeled data contained by each client by assigning communities.

\textbf{Baselines.}
For FGL Optimization, we compare FedGTA with FedAvg~\cite{mcmahan2017fedavg}, FedProx~\cite{li2020fedprox}, Scaffold~\cite{karimireddy2020scaffold}, MOON~\cite{li2021moon}, FedDC~\cite{gao2022feddc}, and GCFL+~\cite{xie2021gcfl}.
For FGL Model, we conduct comparisons on recently proposed FedGL~\cite{chen2021fedgl} and FedSage+~\cite{zhang2021fedsage}. 
For local models, we utilize simple and scalable GCN~\cite{welling2016gcn}, GraphSage~\cite{hamilton2017graphsage}, SGC~\cite{wu2019sgc}, SIGN~\cite{frasca2020sign}, S$^2$GC~\cite{zhu2021ssgc}, GBP~\cite{chen2020gbp}, and GAMLP~\cite{zhang2022gamlp}.
Based on this, the results we present are calculated by 10 runs.
Unless otherwise stated, we adopt GAMLP as the local model and employ all datasets with Louvain 10 clients split, except for ogbn-papers100M, which is divided into 500 clients. 
Notably, we experiment with multiple existing scalable GNN models in separate modules to validate the generalizability of our optimization strategy and avoid complex charts, making the results more reader-friendly.

\textbf{Hyperparameters.}
The hyperparameters in the local model are set according to the original paper if available.
Otherwise, we perform automatic hyperparameter optimization via the Optuna toolkit~\cite{akiba2019optuna}.
Specifically, we explore the optimal values for feature propagation steps ($k$) and model layers ($L$) within the ranges of 2 to 20 and 2 to 6.
Regarding the percentage of selected augmented nodes and the number of generated neighbors, we conduct a grid search from $\{$0.01, 0.05, 0.1, 0.5$\}$ and $\{$2, 5, 10$\}$ respectively.
The hidden dimension for the small dataset is set to 64 with the number of local epochs set to 3. 
For medium or large-scale datasets, we set 256, and 5, respectively.
We default perform 100 rounds, and the coefficient of the gradient regularization terms is determined through a grid search with values $\{$0.001, 0.01, 0.1$\}$.
The optimal window size of gradient dynamic clustering ranges from 2 to 10.
For our proposed FedGTA, the order of moments ($K$) and the similarity threshold ($\epsilon$) are explored within the ranges of 2 to 20 and 0 to 1.


\textbf{Experiment Environment.}
The experiments are conducted on the machine with Intel(R) Xeon(R) Gold 6230R CPU @ 2.10GHz, and NVIDIA GeForce RTX 3090 with 24GB memory and CUDA 11.8.
The operating system is Ubuntu 18.04.6 with 216GB memory.

\subsection{Performance Comparison}
To answer \textbf{Q1}, we report the transductive performance in Table~\ref{tab: trans_cmp}, where FedGTA outperforms other baselines. 
Specifically, compared to the second result, FedGTA achieves an average improvement of 2.33\% and 2.54\% when using GCN and GAMLP. 
While FedGL and FedSage+ outperform methods using GCN as the local model in some cases, they cannot achieve more competitive results and even fail to handle large-scale scenarios due to their limited scalability. 

The experiment results in Table~\ref{tab: ins_exp} show that FedGTA consistently outperforms all the baselines under the inductive setting. 
Compared to the most competitive MOON and FedDC, FedGTA has a lead of more than 3.5\% and 2.2\%, respectively.  
Notably, we only compare FedGTA with other FGL optimization studies (\textbf{bold} or \underline{underline}).

\subsection{Generalization}
To answer \textbf{Q2}, we demonstrate that FedGTA can be applied to a large variety of GNN variants: 
GCN, GAMLP, SIGN, S$^2$GC, SGC, GraphSAGE, and GBP are shown in Table~\ref{tab: trans_cmp}, Table~\ref{tab: ins_exp}, and Table~\ref{tab: ab_exp}.
Building upon the above backbone GNNs, to test the effectiveness of our proposed FedGTA, we evaluate it on both coupled and sampling-based GNN models, which differ in the orderings of feature propagation and transformation.
Through the above experiments, we observe that FedGTA consistently outperforms the other FGL optimization baselines in both GNN categories.

As we claimed, FedGTA is an optimization strategy suitable for FGL, and a natural idea is to combine it with the existing FGL Model studies. 
In Table~\ref{tab: fedmodel_fedgta}, we present the experimental results of combining FedGTA and other competitive strategies with the FGL Model studies.
As shown in the Tabel\ref{tab: fedmodel_fedgta}, the test accuracy of FedGTA could improve FedGL and FedSage+ by an average of more than 2.5\% on three datasets. 
Therefore, we conclude that FedGTA can generalize to different types of GNNs and existing FGL Model studies well.

\begin{table}[]
\setlength{\abovecaptionskip}{0.2cm}
\setlength{\belowcaptionskip}{-0.2cm}
\caption{Inductive performance under 10 clients Metis split.
}
\scriptsize 
\label{tab: ins_exp}
\scalebox{1.5}{
\begin{tabular}{cc|cc}
\midrule[0.3pt]
Model                    & Optimization & Flickr              & Reddit              \\ \midrule[0.3pt]
\multirow{7}{*}{SIGN}    & FedAvg       & 48.10±0.31          & 91.31±0.10          \\
                         & FedProx      & 48.58±0.27          & 91.15±0.12          \\
                         & Scaffold     & 48.98±0.32          & 90.58±0.08          \\
                         & MOON         & {\ul 49.34±0.27}    & 91.37±0.11          \\
                         & FedDC        & 48.76±0.20          & {\ul 91.46±0.05}    \\
                         & GCFL+        & 48.58±0.16          & 90.54±0.07          \\
                         & FedGTA       & \textbf{50.89±0.18} & \textbf{93.79±0.06} \\ \midrule[0.3pt]
\multirow{7}{*}{S$^2$GC} & FedAvg       & 48.75±0.30          & 92.31±0.09          \\
                         & FedProx      & 48.81±0.18          & 92.67±0.07          \\
                         & Scaffold     & 48.65±0.29          & 92.39±0.13          \\
                         & MOON         & {\ul 49.36±0.26}    & 92.65±0.07          \\
                         & FedDC        & 48.91±0.26          & {\ul 93.30±0.12}    \\
                         & GCFL+        & 49.24±0.14          & 93.06±0.03          \\
                         & FedGTA       & \textbf{51.32±0.19} & \textbf{95.07±0.08} \\ \midrule[0.3pt]
\end{tabular}
}
\end{table}

\begin{table}[]
\setlength{\abovecaptionskip}{0.3cm}
\setlength{\belowcaptionskip}{-0.2cm}
\caption{Performance gain in FGL Model under 10 clients Metis split.
}
\footnotesize 
\label{tab: fedmodel_fedgta}
\resizebox{\linewidth}{20mm}{
\setlength{\tabcolsep}{1.5mm}{
\begin{tabular}{cc|ccc}
\midrule[0.3pt]
Model                     & Optimization & ogbn-arxiv        & Flickr            & Reddit            \\ \midrule[0.3pt]
\multirow{4}{*}{FedGL}    & FedAvg       & 70.5±1.2          & 47.9±0.3          & 89.1±0.3          \\
                          & MOON         & 70.9±1.4          & 48.1±0.4          & 88.7±0.3          \\
                          & FedDC        & 70.3±1.9          & 47.8±0.5          & 89.3±0.5          \\
                          & FedGTA       & \textbf{72.3±0.9} & \textbf{50.5±0.3} & \textbf{92.0±0.2} \\ \midrule[0.3pt]
\multirow{4}{*}{FedSage+} & FedAvg       & 69.8±1.9          & 48.1±0.5          & 90.2±0.3          \\
                          & MOON         & 70.4±1.7          & 48.5±0.5          & 90.2±0.3          \\
                          & FedDC        & 69.8±2.1          & 48.2±0.7          & 90.4±0.5          \\
                          & FedGTA       & \textbf{72.5±1.4} & \textbf{51.4±0.4} & \textbf{92.9±0.3} \\ \midrule[0.3pt]
\end{tabular}
}}
\end{table}

\begin{table}[]
\setlength{\abovecaptionskip}{0.2cm}
\setlength{\belowcaptionskip}{-0.2cm}
\caption{Ablation study on three scalable GNN models.
}
\footnotesize 
\label{tab: ab_exp}
\resizebox{\linewidth}{25mm}{
\setlength{\tabcolsep}{1.2mm}{
\begin{tabular}{cc|cccc}
\midrule[0.3pt]
\multirow{2}{*}{Model}     & \multirow{2}{*}{Component} & \multicolumn{2}{c}{ogbn-products} & \multicolumn{2}{c}{Reddit} \\
                           &                             & Louvain         & Metis           & Louvain      & Metis       \\ \midrule[0.3pt]
\multirow{3}{*}{SGC}       & w/o Mom.                  & 72.9±0.3      & 71.8±0.3      & 91.4±0.1   & 91.9±0.1  \\
                           & w/o Conf.              & 73.6±0.2      & 73.1±0.1      & 92.5±0.1   & 92.6±0.1  \\
                           & FedGTA                      & \textbf{74.2±0.1}      & \textbf{73.6±0.2}      & \textbf{93.0±0.1}   & \textbf{93.1±0.1}  \\ \midrule[0.3pt]
\multirow{3}{*}{GBP}       & w/o Mom.                  & 77.1±0.5      & 77.7±0.5      & 92.4±0.1   & 92.4±0.1  \\
                           & w/o Conf.              & 77.5±0.3      & 78.1±0.4      & 93.3±0.1   & 93.0±0.1  \\
                           & FedGTA                      & \textbf{78.2±0.3}      & \textbf{78.7±0.4}      & \textbf{93.7±0.1}   & \textbf{93.4±0.1}  \\ \midrule[0.3pt]
\multirow{3}{*}{SAGE} & w/o Mom.                  & 73.8±0.5      & 73.1±0.3      & 88.4±0.1   & 88.0±0.1  \\
                           & w/o Conf.              & 75.5±0.2      & 75.6±0.3      & 90.2±0.1   & 89.9±0.1  \\
                           & FedGTA                      & \textbf{76.8±0.1}      & \textbf{76.2±0.3}      & \textbf{90.7±0.1}   & \textbf{90.2±0.1}  \\ \midrule[0.3pt]
\end{tabular}
}}
\end{table}

\begin{figure}[htbp]   
	\centering
    \setlength{\abovecaptionskip}{0.2cm}
    \setlength{\belowcaptionskip}{-0.15cm}
	\includegraphics[width=\linewidth,scale=1.00]{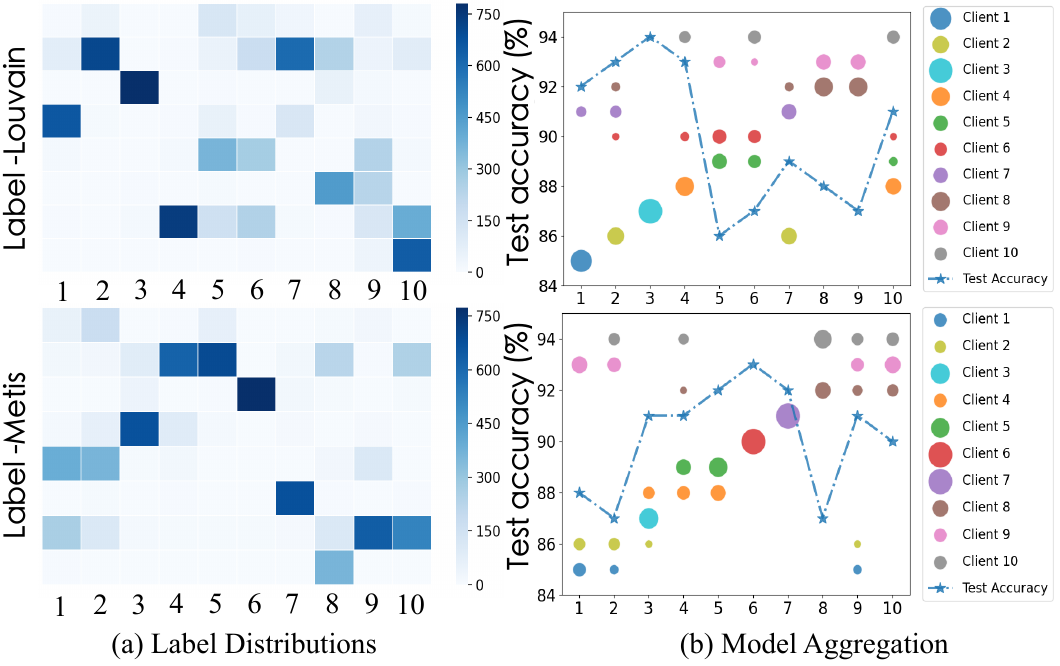}
	\caption{
    The visualization of model aggregation in Amazon Photo with 10 clients split, which contains 8 different node labels.
    The circle size corresponds to aggregation weight, from large to small.
    }
    \label{fig: model_aggregation}
\end{figure}

\begin{figure*}[htbp]   
	\centering
    \setlength{\abovecaptionskip}{0.2cm}
	\includegraphics[width=\linewidth,scale=1.00]{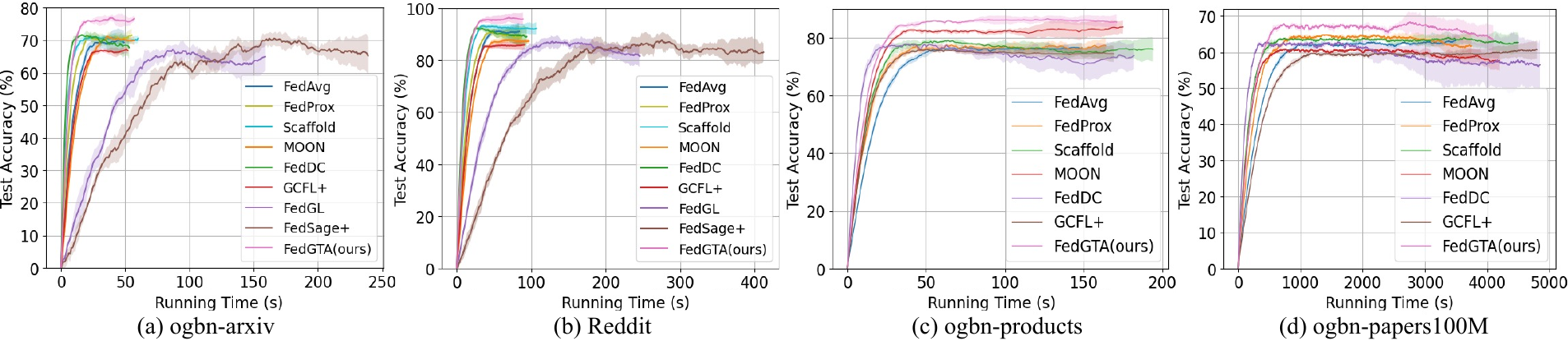}
	\caption{
    Convergence curves of our proposed FedGTA and baseline models on $4$ large-scale benchmark graph datasets. 
    Curves represent the local model training phase and model aggregation phase on the server side. 
    The shaded area is the result range of multiple runs.}
	\label{fig: efficient_scalable_convergence_curves}
\end{figure*}

\begin{figure}[htbp]   
	\centering
    \setlength{\abovecaptionskip}{0.15cm}
    \setlength{\belowcaptionskip}{-0.3cm}
	\includegraphics[width=\linewidth,scale=1.00]{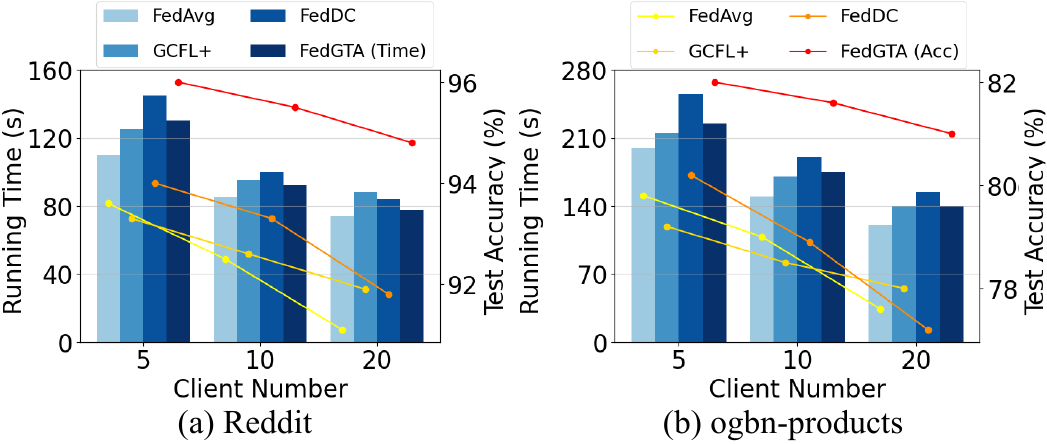}
	\caption{
     Training efficiency with different numbers of clients.
    }
    \label{fig: multi_client}
\end{figure}

\begin{figure}[htbp]   
	\centering
    \setlength{\abovecaptionskip}{0.15cm}
    \setlength{\belowcaptionskip}{-0.3cm}
	\includegraphics[width=\linewidth,scale=1.00]{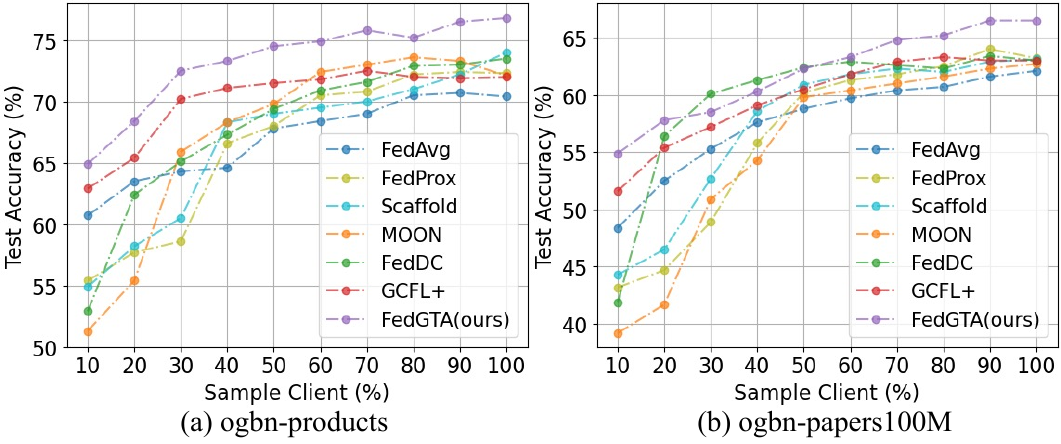}
	\caption{
     Performance with different clients participating.
    }
    \label{fig: sample_client}
\end{figure}

\subsection{Method Interpretability}
To answer \textbf{Q3}, we focus on the implementation of FedGTA on the client and server side.
Specifically, we present the ablation experiments shown in Table~\ref{tab: ab_exp} to investigate the contribution of local smoothing confidence and mixed moments of neighbor features computed on the client side.
Subsequently, their efficacy hinges on the server's ability to perform efficient model aggregation, thereby achieving optimized federated training.
The visualization of server-side model aggregation based on our method is presented in Fig.~\ref{fig: model_aggregation}.

In the ablation study, we use "Mom." and "Conf." to represent the mixed moments of neighbor features and local smoothing confidence. 
We also use "SAGE" to refer to the GraphSage model.
According to the experimental results shown in Table~\ref{tab: ab_exp}, we observe that both Mom. and Conf. contribute to the improvement of the optimization process significantly.
Furthermore, the combination of them reduces the variance during the federated training process.

In the visualization part, we analyze the model aggregation process by referring to the label distributions of each client as shown in Fig.~\ref{fig: model_aggregation}. 
Based on our design, we aim to achieve personalized aggregation for each client only combines with those with similar label distribution while the smoothness of each subgraph determines the aggregation weight. 
Clearly, for the local subgraph, the steeper the label distribution, the more likely it has a smooth topology. 
This is because connected nodes are more likely to have similar feature distributions or the same label.
In Fig.~\ref{fig: model_aggregation}(b), we select the best aggregation round for presentation.
According to the circle categories in multi-clients, we observe that FedGTA successfully customizes the model aggregation targets for each client. 
To further clarify, the circle sizes of clients reflect both their local smoothing confidence and corresponding aggregation weights. 
This demonstrates the efficiency of our method, as subgraphs with more smoothing play a dominant role in the model aggregation process.

\subsection{Efficiency and Scalability Analysis}
\label{sec:Efficiency and Scalability Analysis}

To answer \textbf{Q4}, we report the running time in Fig.~\ref{fig: efficient_scalable_convergence_curves} and Fig.~\ref{fig: multi_client}, which includes both local training and model aggregation on the server.
Notably, FedGL and FedSage+ consume an extended amount of running time due to their complex local model architectures and additional cross-client interactions.
In contrast, FGL optimization strategies based on scalable GNNs are more efficient. 
Among these optimization studies, FedGTA exhibits the most stable and superior performance as evidenced by the curve and shaded trends in Fig.~\ref{fig: efficient_scalable_convergence_curves}.

According to Table~\ref{tab: algorithm_analysis}, the time complexity of FedGTA is mainly related to the model-agnostic sparse matrix multiplication of soft labels ($c\ll f$): $O(k^2mnc)$ and $O(NkKc)$, which is not dependent on the local training process. 
In contrast, in the client-side model training, FedProx and Scaffold introduce a gradient regularization term $O(f^2)$, MOON and FedDC rely on the outputs from different federated rounds to construct additional model-contrastive loss $O(2nf)$ and model-drift loss $O(4f^2)$. 
Due to this reason, as the local data scale increases and the models become complex, the training cost becomes larger, leading to an increase in the time complexity and unstable performance of the aforementioned optimization methods.
This is validated in Fig.~\ref{fig: efficient_scalable_convergence_curves} and Fig.~\ref{fig: multi_client} of our study.
On the server side, Scaffold introduces additional time complexity of $O(Nf^2+f)$ for updating global control variables. 
The time complexity term of GCFL+ $O((N^2(\log(N)+T^2 f^2))$ becomes sensitive to the window size $T$ used for dynamic gradient clustering and the number of participating clients $N$.
The above inference is also confirmed with Fig.~\ref{fig: multi_client}.
Although FGL optimization strategies, including our methods, share similar inference efficiency, we include the additional experiment discussing the differences in inference efficiency between various GNN models on the ogbn-arxiv dataset with 10-clients Louvain split(time reported in second). 
For FGL Model studies, FedGL(1.10±0.12) and FedSage(1.73±0.18) proved to be the least efficient and unapplicable for solving scalability. 
While decoupled GNN models such as SGC(0.12±0.03), SIGN(0.19±0.07), and GAMLP(0.25±0.08) demonstrate less time cost.

In practical scenarios of FGL, there are often a large number of clients, which makes it necessary to select a subset of clients to participate in each round to reduce communication and time-space costs.
This amounts to performing Louvain 50 clients split for ogbn-products and 500 clients split for ogbn-papers100M. 
In Fig.~\ref{fig: sample_client}, we present the experimental results. 
According to the experimental results, we conclude that the accuracy level of model and embeddings comparison-based approaches, such as MOON and FedDC, significantly dropped due to the high heterogeneity of subgraphs when the participation ratio is small.
In contrast, personalized strategies, such as FedGTA and GCFL+, exhibit robustness. 
However, compared with FedGTA, the implicit utilization of structural topology in GCFL+ causes its inability to produce competitive performance.

\section{Conclusion}
This paper presents the first integration of large-scale graph learning with FGL, motivated by the need for analyzing real-world applications.
Large-scale graph learning can be computationally intensive and space-consuming, which can be effectively solved with FGL due to its decentralized structure. 
In this paper, we first discuss the flaws of existing FGL approaches. 
Specifically, FGL Model studies lack scalability due to complex models, and most FGL optimization strategies adopted by FGL fail to recognize the topology.
To address the above issues, we propose FedGTA, which is the first topology-aware optimization strategy for FGL. 
Experimental results demonstrate that FedGTA significantly outperforms competitive baselines in terms of model performance and generalizability.

FedGTA incorporates Non-param LP, which allows for the explicit consideration of both model prediction and topology in each client. 
This strategy is straightforward and user-friendly. 
However, a promising avenue for improvement is to leverage additional information provided by local models during training, such as $k$-layer propagated features. 
Moreover, we employ personalized model aggregation based on mixed moments of neighbor features' similarity, which has shown effectiveness in a data-driven context. 
Nevertheless, there is potential for exploring an adaptive aggregation mechanism that considers the impact of topology on the FGL.

\begin{acks}
 This work was partially supported by 
 (i) the National Key Research and Development Program of China 2021YFB3301301, 
 (ii) NSFC Grants U2241211, 62072034, and 
 (iii) CCF-Huawei Populus Grove Fund. Rong-Hua Li is the corresponding author of this paper.
\end{acks}

\newpage
\bibliographystyle{ACM-Reference-Format}
\balance
{
\bibliography{sample}
}

\end{document}